\documentclass{article} 
\usepackage{paperstyle,times}
\usepackage{graphicx}

\usepackage{float} 

\usepackage{amsmath,amsfonts,bm}

\def\eqref#1{equation~\ref{#1}}

\def\1{\bm{1}}

\DeclareMathAlphabet{\mathsfit}{\encodingdefault}{\sfdefault}{m}{sl}
\SetMathAlphabet{\mathsfit}{bold}{\encodingdefault}{\sfdefault}{bx}{n}

\usepackage{hyperref}
\usepackage{url}

\title{Interpreting and Steering Protein Language Models through Sparse Autoencoders}

\author{%
  \phantom{textte}
  \And
  Edith N. Villegas Garcia 
  \And
  Alessio Ansuini
  \And
  \phantom{texttext}
  \AND 
  \hspace{4.2cm} Area Science Park, Trieste, Italy\\
  \hspace{3.1cm} \texttt{\{edith.villegas, alessio.ansuini\}}\\  
  \hspace{4.7cm} \texttt{@areasciencepark.it}}

\lmrlfinalcopy 

\begin{document}

\maketitle

\begin{abstract}

The rapid advancements in transformer-based language models have revolutionized natural language processing, yet understanding the internal mechanisms of these models remains a significant challenge. This paper explores the application of sparse autoencoders (SAE) to interpret the internal representations of protein language models, specifically focusing on the ESM-2 8M parameter model. By performing a statistical analysis on each latent component’s relevance to distinct protein annotations, we identify potential interpretations linked to various protein characteristics, including transmembrane regions, binding sites, and specialized motifs. We then leverage these insights to guide sequence generation, shortlisting the relevant latent components that can steer the model towards desired targets such as zinc finger domains. This work contributes to the emerging field of mechanistic interpretability in biological sequence models, offering new perspectives on model steering for sequence design.
\end{abstract}

\section{Introduction}\label{introduction}

Since the introduction of the transformer architecture \citep{vaswani2017attention}, the capabilities of neural networks to model and generate natural language have increased dramatically. Yet, due to their black-box nature, we still lack a clear understanding of how these models achieve such capabilities \citep{rai2024practical}. Recently, the mechanistic interpretability approach has been proposed, where researchers try to reverse engineer neural networks in a way similar to reverse engineering computer programs \citep{chris2022mecin, rai2024practical}. This involves understanding which features the network is learning from our input data, and then how it performs operations with this set of features. 

It has been observed that neural networks tend to encode high-level features as linear directions in their representation space—such as the gender direction in word embeddings \citep{park2023linear}. Additionally, these models can store more facts and features than their parameter counts would suggest, a phenomenon known as superposition \citep{elhage2022superposition}. This phenomenon represents a core problem for interpretability: as a single neuron activation can be polysemantic and represent multiple features simultaneously.

Recently, sparse autoencoders (SAEs) have been proposed as a method to disentangle internal representations in language models, extracting features from superposition in an unsupervised manner \citep{claude3sonnet, bricken2023towards, cunningham2023sparse}. Notably, these features appear to be actionable: artificially activating them during inference can steer a model’s output \citep{claude3sonnet, makelov2024sparse}. Such methods have been successfully applied to language \citep{claude3sonnet, cunningham2023sparse, gao2024scaling}, as well as vision and multimodal models \citep{gorton2024missing, surkov2024unpacking}, but biological and protein sequence models remain relatively unexplored \citep{simon2024interplm, adamsmechanistic}.

Protein language models have been shown to encode structural, functional, and evolutionary information in their internal representations \citep{rives2019biological, lin2022language, hayes2024simulating}. Interpretability methods for these models could reveal biological mechanisms, and support model debugging and editing for safety considerations. Additionally, model steering can be incorporated into sequence design pipelines.

The main contributions of this paper are:

• A trained sparse autoencoder (SAE) for the ESM-2 8M parameter model, along with potential interpretations of its latent components (sections \ref{sec:saedef}, \ref{sec:trainingmain} and \ref{interpretinglatents}).

• A methodology for generating protein sequences by intervening on specific latents, demonstrating successful steering towards non-trivial features, such as zinc finger domains (section \ref{sec:steeredseqs}).

• A heuristic for selecting the model layer from which to extract representations using an intrinsic dimension estimator (section \ref{sec:layerselection}).

\section{Background}

\subsection{Protein Language Models} 

Many advancements in Natural Language Processing have been successfully applied to biological sequence modeling. Transformer-based neural networks can be trained on protein sequences using the Masked Language Modeling (MLM) task, where each amino acid is treated as a token that can be randomly masked. The model learns to predict the masked tokens by minimizing the following loss function \citep{rives2019biological}:

\begin{equation}
\mathcal{L}_\mathrm{MLM} = \mathbb{E}_{x \sim X}  \mathbb{E}_{M} \sum_{i \in M} - \mathrm{log}   \, p(x_i | x_{/M} )
\end{equation}

where $x$ is a protein sequence, $M$ is a set of masked indices and $p(x_i | x_{/M} )$ is the probability assigned to the ground truth amino acid $x_i$ given its sequence context. 

Training on the Masked Language Modeling (MLM) task forces the network to learn dependencies between masked amino acids and their sequence context while simultaneously capturing various biological features present in the data. Embeddings extracted from these models have been shown to encode information about secondary structure, tertiary contacts (residue-residue interactions), function, remote evolutionary relationships, and factors relevant to predicting mutational effects \citep{rives2019biological, elnaggar2021prottrans, meier2021language, lin2022language, hayes2024simulating}. On the other hand, the attention mechanism appears to prioritize binding sites, with attention maps capturing information about residue-residue interactions. \citep{vig2020bertology}. 

\subsection{Sparse Autoencoders} \label{sec:saedef}

The sparse autoencoders used for interpretability are simple, single-layer models trained on the activations of a larger language model. To disentangle network features, the hidden layer is made significantly larger than the original embeddings, creating an overcomplete basis. A sparsity constraint is then applied to ensure that only a few latent neurons are active at a time, making the SAE’s hidden representation far more interpretable than standard language model components \citep{claude3sonnet, bricken2023towards, cunningham2023sparse}.

\subsubsection{Architecture}\label{architecture}

The autoencoder is composed of an encoding and a decoding function, given by: 

\begin{equation}
 z = f_{enc}(x) = \mathrm{ReLU}(W_{enc}(x-b_{dec})+b_{enc})
\end{equation}
\begin{equation}
\check{x} = f_{dec}(z) = (W_{dec} \cdot z+b_{dec})
\end{equation}

Here $f_{enc}$ is the encoder, that takes an embedded amino acid token $x \in {\rm I\!R}^{d}$ from a given layer in the model and returns a latent $z \in {\rm I\!R}_{\geq 0}^{ n } $ with a hidden dimension $n$ that is $m$ times bigger that of the original vector (expansion factor). The decoder $f_{dec}$ approximately reconstructs $x$ given $z$, through the decoding matrix $W_{dec} \in {\rm I\!R}^{ n \times  d  }$ and the bias weight $b_{dec} \in {\rm I\!R}^{ d }$. 

The loss function used for the training is a combination of the reconstruction error of the autoencoder $\mathcal{L}_{MSE}$ plus a sparsity constraint $\mathcal{L}_{L_1}$:

\begin{equation}
\mathcal{L}(x) = \mathcal{L}_{MSE} + \mathcal{L}_{L_1} = \sum_{d} ( x_d - \check{x}_d )^{2} + \lambda \sum_{n} z_n
\end{equation}

While training, we renormalize the $W_{dec}$ matrix to have unit norm after each backward pass. This is necessary to prevent that autoencoder latents become arbitrarily small and satisfy the $L_1$ constraint without actually being sparse. 

\section{Methods}\label{methods}

\subsection{Training Details}
\label{sec:trainingmain}

We use the ESM-2 family of models \citep{lin2022language} as our base, extracting activations from the final output of the transformer block. We train on approximately 15k non-redundant protein sequences from SCOPe 2.08 \citep{fox2014scope}. Further details on the architecture, training procedures, and hyperparameter selection can be found in section \ref{sec:training} of the appendix.

\subsection{Layer Selection}
\label{sec:layerselection}

We adopt a principled strategy to select the layer from which we extract representations for the sparse autoencoder. The initial intuition, in line with earlier studies \citep{claude3sonnet, gao2024scaling}, is to choose a mid-to-late layer, where the model is assumed to have developed abstract features but is not yet focused on the output reconstruction task. However, unlike these prior works, we move beyond mere intuition by incorporating a quantitative measure based on intrinsic dimension.

Specifically, we compute the intrinsic dimension of each layer’s representations using the estimator proposed by \citep{facco2017estimating}, and then identify where this value plateaus. 
Previous research has shown that layers corresponding to local minima or plateaus in intrinsic dimension are where abstract information is most clearly encoded \citep{val2024geometry}. Selecting a layer within this plateau increases the likelihood of capturing meaningful representations, providing a stronger foundation for interpretability and model steering.

\subsection{Interpreting Autoencoder Latents}\label{interpretinglatents}

We extract protein annotations from the UniProt database \citep{uniprot2025uniprot} and convert them into binary labels for each amino acid in the sequence. We then compute the precision $\pi$ and recall $\rho$ of each latent component $k$ in detecting a given feature $\phi$.

Let $A$ be the set of all amino acids and  $A_{\phi^{+}}$ the set of amino acids that have been annotated with the feature $\phi$. Considering a latent $k$ to be  active ($k^+$) for a given amino acid when its value $z_k$ exceeds a certain threshold $\tau_z$, we have: 

\begin{equation}
\pi = P(\phi^{+}|k^{+}) = \frac{\left| \left\{ a \in A_{\phi^{+}}: z_k > \tau_z \right\} \right|}{\left| \left\{ a \in A: z_k > \tau_z \right\} \right|} 
\end{equation}

\begin{equation}
\rho = P(k^{+}|\phi^{+}) = \frac{\left| \left\{ a \in A_{\phi^{+}}: z_k > \tau_z \right\} \right|}{\left| A_{\phi^{+}} \right|} 
\end{equation}

This gives us a value of precision and recall for each pair of $k$, $\phi$. We consider a latent component to be associated with a specific feature if its precision or recall exceed a predefined threshold that we set to $0.80$.

\subsection{Generating steered sequences}
\label{sec:steeredseqs}

\begin{figure}[h]
\begin{center}
\includegraphics[width=5.0in]{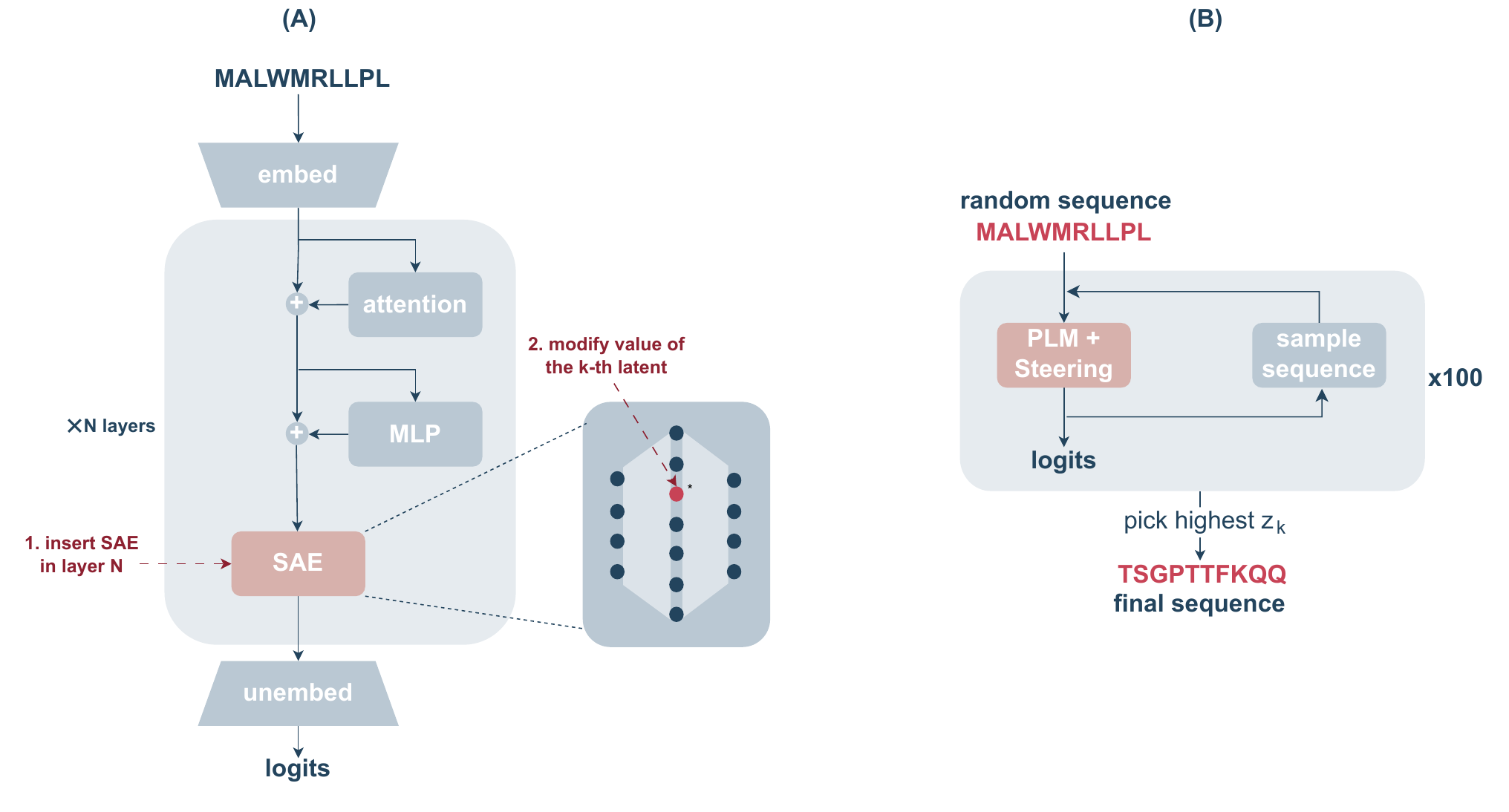} 
\end{center}
\caption{Sequence generation procedure. (A) To steer the model outputs, the base Protein Language Model is modified through the insertion of a sparse autoencoder in the residual stream, at a particular layer. During inference, the value of one of the latents in the autoencoder is modified. (B) Starting from a random sequence, we perform inference with the modified and intervened model, and sample a new sequence from the output logits. We repeat this procedure iteratively a certain number of times (i.e. 100), and at the end we retain the sequence which gives the highest value for the activation of the target latent $z_k$.}
\label{fig:steeringplot}
\end{figure}

Once a latent corresponding to a specific feature is identified, we can steer the model during inference to increase the likelihood of generating protein sequences that contain that feature. This approach, previously demonstrated in natural language models by \citep{claude3sonnet}, is outlined in figure \ref{fig:steeringplot}.
 
We begin with a randomly generated amino acid sequence of fixed length. After a forward pass through the model and the encoder layer of the SAE, we modify the target latent $z_k$ by scaling and shifting its value to increase its magnitude (\eqref{eq:scale}). We then pass the modified value $z_k^*$ through the decoder layer $f_{dec}$ and add back the original reconstruction error of the embedding $x$ before passing it through the rest of the model (\eqref{eq:decode_err}).

\begin{equation}
z_k^* = a \cdot z_k + b
\label{eq:scale}
\end{equation}
\begin{equation}
x^* = f_{dec}(z_k^*)  + x_{err} 
\label{eq:decode_err}
\end{equation}

For each position in the sequence, we randomly sample an amino acid according to the probability distribution predicted by the model under the intervention. We repeat this process starting from the predicted sequence and perform 100 iterations of inference-prediction to refine the sequence. We select the sequence at the iteration where the value of the activation $z_k$ is maximum.

\section{Results}\label{results}

\subsection{Interpreting Latents}

For the interpretability analysis, we focus on the autoencoder that provides the best trade-off between sparsity and reconstruction quality, as described in section \ref{sae_selection} of the appendix. We compute recall and precision for all [$k$, $\phi$] pairs, following the methodology outlined in section \ref{interpretinglatents}, using three increasing thresholds of latent activation. This allows us to assess the robustness of the identified features.

We find 395 putative [$k$, $\phi$] associations, detailed in table \ref{latentfeature-table}. Among these, there are latent components associated to different binding sites, cellular regions and motifs like zinc fingers. The complete set of latent - feature associations is available in the supplementary data (see section \ref{sec:code_data_avail}).

\begin{figure}[h]
\begin{center}
\includegraphics[width=5.5in]{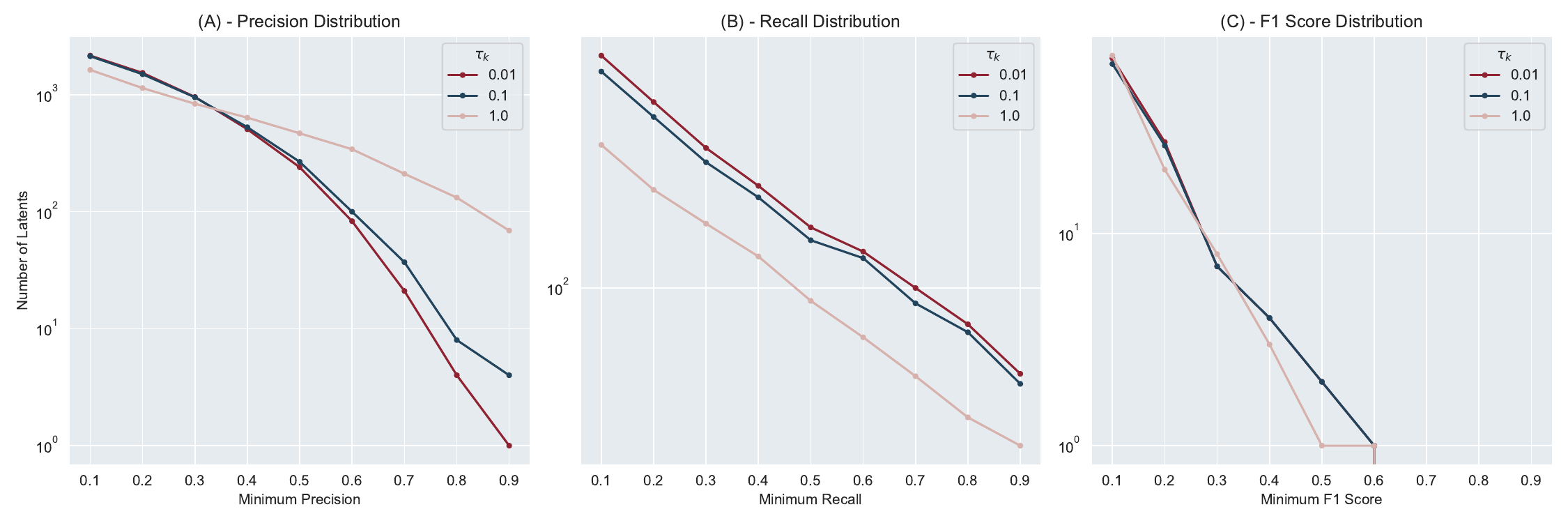}
\end{center}
\caption{Distribution of the number of latent SAE components that detect a feature with a minimum value of precision, recall and F1-score. Setting a higher value for the activation threshold $\tau_k$ significantly increases the precision with which latents detect features, but it decreases the recall.}
\label{fig:distributions}
\end{figure}

We also identify many potential associations with a lower confidence (lower values of precision or recall). To get an idea of how many putative association are found for each value of precision/recall (as well as a combined F1-score) we plot their cumulative distributions in figure \ref{fig:distributions}.

Intuitively, a latent component that perfectly matches an annotation type should exhibit both high precision and recall, resulting in a high F1-score. However, since the model is trained to optimize a masked language modeling loss, the features it learns may not directly align with those in the manually curated dataset. For instance, a latent $k$ might encode a more specific subcategory of a dataset label $\phi$, such as identifying the starting amino acid of a helix rather than the entire helix structure (as seen by \citet{adamsmechanistic} on some features). In such cases, the association between $k$ and $\phi$ would likely have high precision but low recall.

Similarly, a high recall but low precision may indicate that the model has learned a more coarse-grained feature than those defined in the dataset. This is evident in cases such as latent $k=610$, which activates across various types of zinc fingers, and latent $k=555$, which responds to alpha-keto acids. To prevent selecting latents with high recall due to trivial reasons -- such as activating indiscriminately on all amino acids -- we also assess the proportion of times a latent is active on amino acids lacking a given label, denoted as $P(k^+|\phi^-)$, before confirming an association. In section \ref{sec:interpretinglatents}, we present examples of the distributions of $P(k^+|\phi^+)$ and $P(k^+|\phi^-)$
for a zinc finger region.

\begin{table}[t]
\caption{Number of latent-feature annotation pairs with a minimum precision/recall of 0.8 for different values of the activation threshold $\tau_k$}
\label{latentfeature-table}
\begin{center}
\begin{tabular}{cccc}
\multicolumn{1}{c}{\bf $\tau_k$}  & \multicolumn{1}{c}{\bf \# Pairs (Precision \textgreater{} 0.8)} 
& \multicolumn{1}{c}{\bf \# Pairs (Recall \textgreater{} 0.8)} 
& \multicolumn{1}{c}{\bf Total}\\ 
\hline
0.01 & 4 & 262 & 266\\
0.10 & 8 & 234 & 242\\
1.00 & 133 & 61 & 194\\
Total (unique) & 133 & 262 & 395 \\
\end{tabular}
\end{center}
\end{table}

\subsection{Sequence Generation}

\begin{figure}[h]
\begin{center}
\includegraphics[width=6.0in]{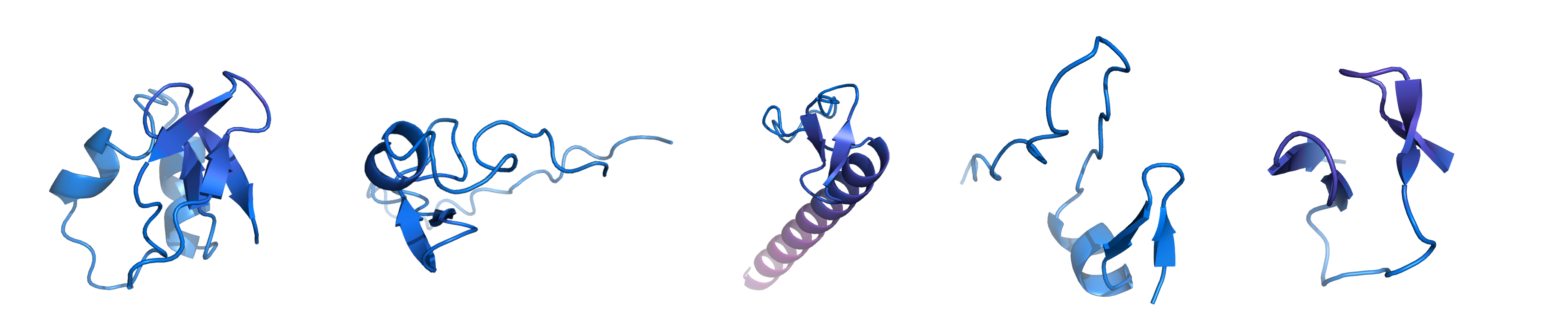} 
\end{center}
\caption{Examples of generated sequences subsequently folded with ESMFold \citep{lin2022language}. The sequences were generated while intervening on the model by increasing the value of latent components associated to the zinc finger motif. With the intervention, the model has a tendency to generate pairs of beta sheets in the vicinity of a helix, as in a typical zinc finger structure.}
\end{figure}

We test the sequence generation procedure with shortlisted latents with a clear association to the zinc finger region annotations (high recall in this specific case). We probe different values of sequence length ($[22, 27, 30, 35, 40, 60]$), of the scaling $a$ ($[2, 5, 10, 20, 30]$) and shift $b$ ($[0.1, 1, 10, 50, 100, 200]$), for a total of 180 combinations, obtaining one sequence for each.

Using an online tool for automatic annotation of zinc finger regions \citep{sathyaseelan2023sequence}, we look for known zinc finger motifs and Pfam family matches in our generated sequences. We find 24 matching regions (out of 180 sequences) when we simultaneously intervene on the two most prominent latents (highest recall) displaying a zinc finger association. In contrast, intervening on the most prominent latent produces only 3 matches, while generating sequences from the baseline model or intervening on a random latent or a random pair of latents produces no matches in any case. Among the matched sequences, the highest percent similarity was $48\%$, with an average sequence similarity of $31\%$, indicating a good level of diversity among the generated sequences.

While this sequence generation pipeline requires parameter fine-tuning to improve the success rate, the process can be automated by introducing appropriate heuristics to search the parameter space efficiently. To the best of our knowledge, this is the first application of steering with sparse autoencoder features to generate complex protein sequences, extending beyond trivial features like specific amino acids or simple amino acid repeats.

\section{Discussion and conclusions}\label{discussion}

In this study, we demonstrate the potential of sparse autoencoders (SAEs) for interpreting and manipulating the internal representations of protein language models. By training a SAE on the ESM-2 8M parameter model, we identified and interpreted latent features associated with various protein annotations, including transmembrane regions, binding sites, and zinc finger motifs.

We have also demonstrated, for the first time, that these latent components can be leveraged to steer the model towards generating protein sequences with non-trivial structural features, like zinc finger motifs. The results highlight the utility of SAEs in disentangling the complex, polysemantic representations within protein language models, paving the way for more interpretable and controllable sequence generation. This approach not only deepens our understanding of how these models encode biological features but also opens up new possibilities for protein design and engineering. Future work could extend these methods to larger models and a wider range of protein features, further bridging the gap between interpretability and practical applications in computational biology.

\subsubsection*{Code \& Data Availability}\label{sec:code_data_avail}
The weights for the trained sparse autoencoder model are available from huggingface at: \href{https://huggingface.co/evillegasgarcia/sae_esm2_6_l3}{https://huggingface.co/evillegasgarcia/sae\_esm2\_6\_l3}. 
The code is available from github at: \href{https://github.com/edithvillegas/plm-sae}{https://github.com/edithvillegas/plm-sae}. 
Supplementary data is available from zenodo at \href{https://doi.org/10.5281/zenodo.14837817}{https://doi.org/10.5281/zenodo.14837817} 

\subsubsection*{Acknowledgments and disclosure of funding}

We thank the technical support of the Laboratory of Data Engineering staff and acknowledge the AREA Science Park supercomputing platform ORFEO.

A.A. ws supported by the project “Supporto alla diagnosi di malattie rare tramite l'intelligenza artificiale"- CUP: F53C22001770002, and by the European Union – NextGenerationEU within the project PNRR "PRP@CERIC" IR0000028 - Mission 4 Component 2 Investment 3.1 Action 3.1.1. E.N.V.G. was supported by the project PON “BIO Open Lab (BOL) - Raforzamento del capitale umano”—CUP: J72F20000940007.

\bibliography{references}
\bibliographystyle{bibstyle}

\appendix
\section{Appendix}

\subsection{Sparse Autoencoder Training}\label{sec:training}

\subsubsection{Training Dataset}\label{subsec:training_data}

We train our model using the Astral SCOPe 2.08 dataset, filtered to $40\%$ sequence identity, which includes approximately 15k highly non-redundant protein sequences \citep{fox2014scope}. This dataset provides a manageable number of tokens, enabling faster iteration over different hyperparameters while maintaining a diverse range of protein sequences and structural domains. This diversity allows the autoencoder to learn a broad spectrum of features. 

\subsubsection{Handling of dead latents }\label{subsec:dead_latents}

We check how frequently each of the latents activates over a subsample of tokens (50 batches of 4096 tokens) at regular intervals during training (every 500 batches). When this frequency is close to zero ($<10^{-5}$), we consider that the latent is ``dead'' and we re-initialize its weights to ``revive'' it.

\subsubsection{Evaluation Metrics} \label{subsec:eval_metrics}

To decide which hyperparameters (learning rate, sparsity penalty $\lambda$, SAE hidden size $n$) produce the best sparse autoencoder, we use the following metrics:
\begin{itemize}
\item $L_0$: The average number of non-zero components in the latent vector $z$ for a given amino acid token. This is our measure of the sparsity level of the autoencoder. 
\item Number of dead latents: The number of components in the latent space that are never non-zero over a large number of sample tokens ($\sim 10^{5}$). This is a general metric for sparse autoencoders quality. 
\item Cross Entropy (CE) Increase: Difference between the average cross entropy loss of the original model and the cross entropy loss of the model when we substitute the activations in a given layer by the corresponding activations reconstructed by the autoencoder. This indicates how much of the model’s performance the sparse autoencoder fails to reconstruct. 
\end{itemize}

\subsubsection{Hyperparameter selection} \label{subsec:hyperparameter_selection}
We perform a hyperparameter sweep with the following values: learning rate [$5e^{-4}$, $1e^{-4}$, $1e^{-3}$]; $L_1$ penalty [$0.0003, 0.001, 0.005$] and dictionary size multiplier [$32, 10, 5$]. 

We use the evaluation metrics detailed in section \ref{subsec:eval_metrics} to decide on the best combination of hyperparameters. Specifically, we aim to find a model that balances reconstruction error and sparsity by focusing on the ``elbow'' part of the CE increase against $L_0$ plot, where increasing the density of active latents in the latent space does not significantly reduce the CE increase, indicating an optimal trade-off.

\subsection{Interpretability}
\subsubsection{Feature Annotation Data}\label{subsec:annotation_data}
As ground truth features for the interpretability analysis, we use the following protein annotations from Uniprot version 2024\_1 : 
\begin{itemize}
    \item Transmembrane region
    \item Topological domain
    \item Binding site
    \item Zinc finger region
    \item Region of interest
    \item Intramembrane region
    \item Active site
    \item Disulfide bond
    \item Glycosylation site
    \item Helix
    \item Turn
    \item Strand
\end{itemize}

\subsection{Additional Results}
\subsubsection{Intrinsic Dimension Analysis}\label{id_layers}

\begin{figure}[H]
\begin{center}
\includegraphics[width=3.0in]{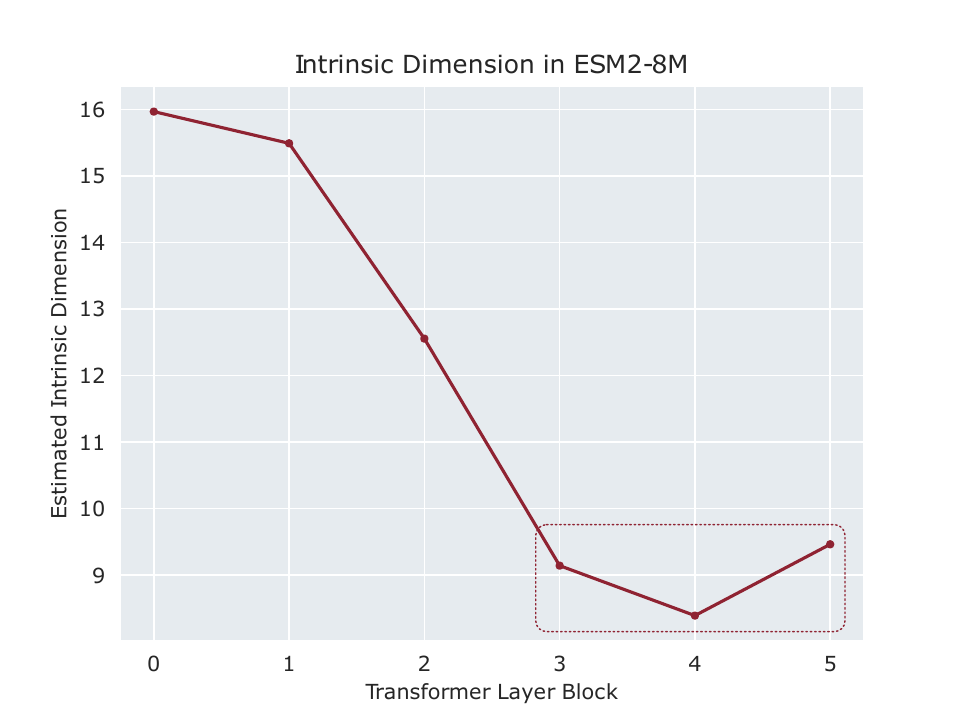}
\end{center}
\caption{Evolution of the intrinsic dimension estimate through the layers of the ESM-2 8M model. We highlight the layers in the plateau/final ascent region.}
\label{fig:idplot}
\end{figure}

We estimate the intrinsic dimension across all layers of ESM-2 8M following the methodology outlined by \citep{val2024geometry}. The resulting curve, shown in Figure \ref{fig:idplot}, guides our decision to extract embeddings from layer 3. Larger models in the ESM-2 family exhibit a similar intrinsic dimension profile across different model sizes, with a more pronounced plateau region of comparatively low intrinsic dimension in the same relative layer position.

\subsubsection{Sparse Autoencoder selection}\label{sae_selection}

\begin{figure}[H]
\begin{center}
\includegraphics[width=3.0in]{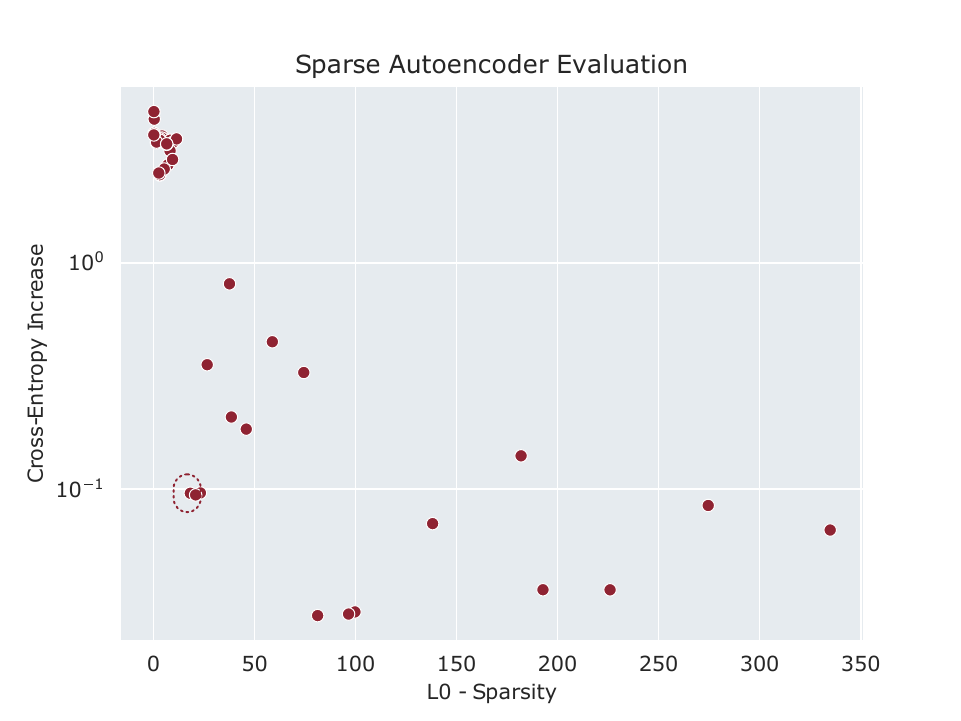} 
\end{center}
\caption{Cross-entropy increase vs sparsity trade-off for all the vanilla sparse autoencoders trained on layer 3 embeddings from ESM-2 8M. The selected autoencoder is indicated by a dashed circle.}
\label{fig:saeval}
\end{figure}

We evaluate all versions of the trained autoencoder primarily on two metrics: cross-entropy increase and sparsity (measured by $L_0$). These two goals are in conflict with each other, so we select what we think is a good compromise at the bend of the Pareto frontier (figure \ref{fig:saeval}). The selected  autoencoder has an average $L_0$ per amino acid of 18, and a cross-entropy increase of 0.10, with a hidden size that is 10 times larger than the original hidden size of the ESM-2 model. The number of dead latents in this SAE is 573. Decreasing the cross-entropy even more would entail a significant increase in activation density, which we want to avoid. 

\subsubsection{Interpreting Latents - Activation Plots}\label{sec:interpretinglatents}

\begin{figure}[H]
\begin{center}
\includegraphics[width=5.5in]{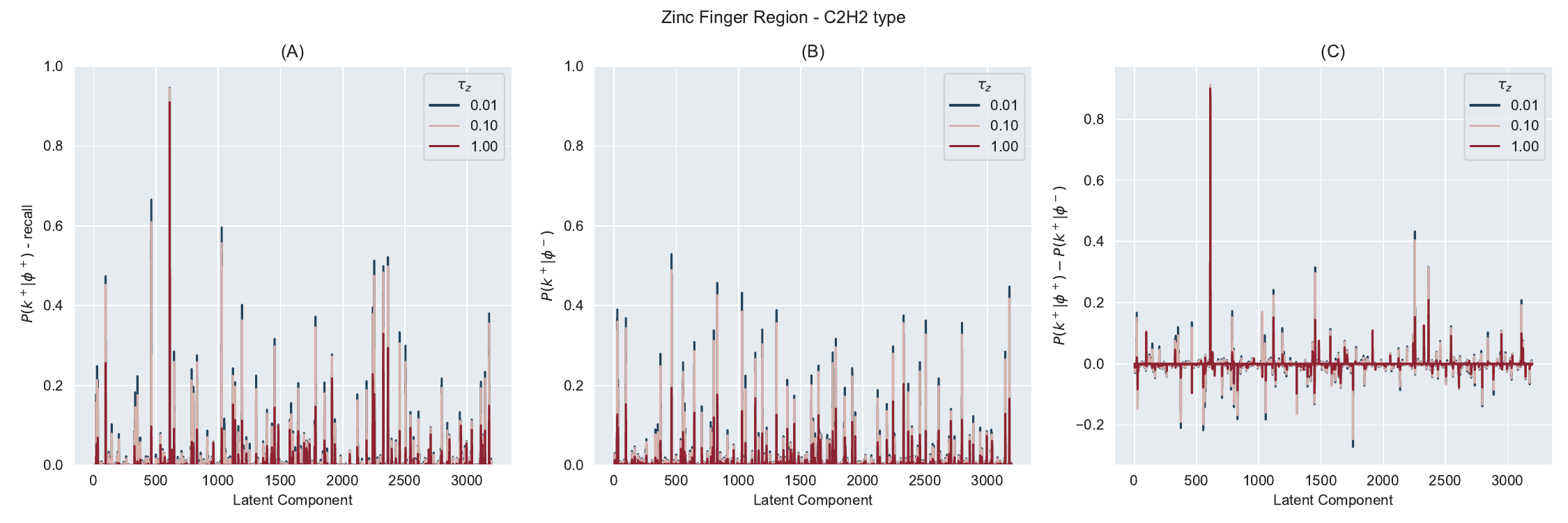}
\end{center}
\caption{
(A) $P(k^+|\phi^+)$ - Percentage of tokens for which each latent component is active when there is a C2H2 zinc finger type label, (B) $P(k^+|\phi^-)$ - percentage when there is no C2H2 zinc finger label, and (C) difference between these two values  for three different activation thresholds $\tau_k$. From the last plot, we see that there is a latent component that is prominently associated with the C2H2 zinc finger label.}
\label{fig:zincfinger}
\end{figure}

\end{document}